\documentclass[letterpaper]{article} 
\usepackage{aaai2026}  
\usepackage{times}  
\usepackage{helvet}  
\usepackage{courier}  
\usepackage[hyphens]{url}  
\usepackage{graphicx} 
\usepackage{natbib}  
\usepackage{caption} 
\usepackage{algorithm}

\usepackage[utf8]{inputenc}
\usepackage{amsmath}
\usepackage{amsthm}
\usepackage{booktabs}
\usepackage{algpseudocode}
\newtheorem{theorem}{Theorem}[section]
\theoremstyle{definition}
\newtheorem{definition}[theorem]{Definition}
\usepackage[switch]{lineno}
\usepackage[font=small,skip=4pt]{caption}
\usepackage[textsize=scriptsize,textwidth=1cm]{todonotes}

\newcommand{\hide}[1]{}

\newcommand{\variableSet}{V}
\newcommand{\axiomSet}{A}
\newcommand{\operatorSet}{O}
\newcommand{\initial}{I}
\newcommand{\goal}{G}
\newcommand{\foda}{\Pi = \langle \variableSet, \axiomSet, \operatorSet, \initial, \goal \rangle}

\newcommand{\basicVariableSet}{V_{b}}
\newcommand{\derivedVariableSet}{V_{d}}
\newcommand{\basicvar}{v}
\newcommand{\derivedvar}{d}

\newcommand{\basicvarDomain}{D_{\basicvar}}
\newcommand{\derivedvarDomain}{\{0, 1\}}

\newcommand{\state}{s}

\newcommand{\partialstate}{p}
\newcommand{\partialstateprime}{p'}
\newcommand{\stateSet}{S}
\newcommand{\partialStateSet}{P}

\newcommand{\cond}[1]{cond_{#1}}
\newcommand{\axiom}{a}

\newcommand{\axiomstuple}{\langle \cond{\axiom}, \derivedvar \rangle}

\newcommand{\operator}{o}
\newcommand{\pre}[1]{pre_{#1}}
\newcommand{\eff}[1]{\mathit{eff_{#1}}}
\newcommand{\operatortuple}{\langle \pre{\operator}, \eff{\operator} \rangle}
\newcommand{\efftuple}{\langle v, d \rangle}

\newcommand{\ps}[1]{p_{#1}}

\newcommand{\prog}{Prog(\ps, \operator)}

\newcommand{\regrapp}{Regr'(\phi, \operator)}
\newcommand{\unt}{Support(\state, \operator)}
\newcommand{\untparam}[2]{Support(#1,#2)}

\usepackage{tikz}

\usetikzlibrary{matrix}
\usetikzlibrary{shapes.geometric} 

\usepackage{newfloat}
\usepackage{listings}
\DeclareCaptionStyle{ruled}{labelfont=normalfont,labelsep=colon,strut=off} 
\floatstyle{ruled}
\newfloat{listing}{tb}{lst}{}
\floatname{listing}{Listing}
\title{Logical Regression for Planning with Axioms}
\author{
    Connor Little,
    Christian Muise
}
\affiliations{%
$\lbrace$ connor.little, christian.muise $\rbrace$ @queensu.ca \\
  Queen's University\\
  Kingston ON, Canada
}

\begin{document}

\maketitle
\begin{abstract}
In automated planning, logical regression is an operation that returns the most general condition necessary for an action to achieve a particular formula. It has many applications, such as allowing for more robust plan execution and providing compact policies for non-deterministic planning. Although relatively simple to calculate in basic planning settings, logical regression becomes significantly more complex when additional factors, such as axioms, are present. We introduce a methodology for approximating the logical regression of an action in a domain that includes axioms; an approximation that limits conditions to partial states. Our method produces minimal partial states while avoiding the recalculation of axioms. To demonstrate the impact of our methods, we embed our form of regression in an execution monitoring context, a well-established setting that can benefit greatly from logical regression. Our results show that this form of regression can dramatically generalize partial states across multiple domains, reducing the number of variables considered for execution monitoring by up to 70\%, and demonstrate that the resulting execution monitor is robust enough to recover frequently in an environment with unexpected changes: several domains recover over 50\% of the time in our tests.
\end{abstract}

\section{Introduction}


In classical planning, a planner begins with a complete initial state and one or more goal conditions. A plan is created by finding a sequence of actions from the initial state to a goal state. The aim of a planner is to generate a plan given a description of a domain. Many powerful language features and techniques have been proposed to help improve planner capabilities.
One such language feature is the use of axioms). Axioms are rules that define the conditions in which a derived variable would be true. This distinguishes these variables from basic variables, which are assigned values through the application of operators. Axioms offer multiple benefits, such as allowing for more natural expressions of phenomena, and a jump in complexity that cannot be efficiently compiled away \cite{Thiébaux_Hoffmann_Nebel_2005}. Axioms facilitate a smaller search space when exploring through best-first search algorithms and can be computed recursively. Despite this, many planners do not support them due to the complexity and additional implementation work they would entail.


While the expressive power of axioms has been explored in the context of plan generation, other related settings in automated planning have had far less focus. We are specifically interested in bringing this expressive power to \textit{logical regression}, a technique typically used for generalized execution monitoring \cite{Fritz_McIlraith} or richer planning settings \cite{muise2024prp}. Logical regression is the calculation of the most general condition that ensures that an action achieves a given formula. Incorporating regression allows some planners, notably non-deterministic planners such as PR2, to become more generalized and more robust by permitting operators to be called on more than a single state. This can be useful when the environment in which the plan will be executed is unpredictable. If, during execution, the state changes unexpectedly, the agent must decide how to recover. Under these conditions, executing the plan exactly as given can lead to unexpected states and inevitably fail.


Calculation of the complete regression formula is complex and expensive in many cases. As such, we introduce an approximation of this regression. The complete regression provides a formula that is biconditional: A formula will hold given a state after the execution of an action if and only if the regression of that formula over an action is entailed by that state. The approximation is less restrictive. It merely states that the formula will hold if the operator is applied. The form of approximation we consider is to restrict the regression to only be represented by partial states.


We introduce a novel method to efficiently compute an approximate logical regression over actions in the presence of axioms. We look at a specialized case of regression in the form of partial states, as it maintains a computationally efficient representation for repeated regression. Our algorithm offers three different approaches for calculating the approximation with varying levels of granularity, from naive to a search based method. Additionally, we look into an exhaustive method and  with full regression in the Appendix.


To explore the benefits that this provides, we apply these methods to an execution monitoring framework. We found that in generating approximations of logical regression, we are able to greatly reduce the number of defined basic variables required to execute a plan in the presence of axioms. The algorithm was able to generalize plans across all domains, with some domains being able to generalize states over 70\% during a plan's execution. We look to demonstrate that the execution monitor can recover in many scenarios without replanning. Overall, we offer three different approaches to approximating regression under the presence of axioms and explore the benefits of each method.

\section{Related Work}



Approximating regression has practical applications when designing planners. PRP and PR2 are examples \cite{Muise_McIlraith_Belle_2014,muise2024prp}. The authors restrict their regression approximation algorithm to only produce partial states and, as such, find the minimal information a state must have to apply the next action. This is useful in the context of the algorithm they use to solve Fully Observable Non-Deterministic (FOND) problems, as it allows partial solutions to generalize to many more explicit states than those explored. This form of regression is an approximation in the presence of conditional effects and is referred to as PRIMF (PRe-IMage Filter) \cite{Muise_McIlraith_Belle_2014}.



Regression has also been used in generalized planning to create state-of-the-art solvers by taking advantage of the generalization it enables. \citeauthor{Chen_Hofmann_Klassen_McIlraith} introduce MOOSE, a generalized planner that uses goal regression to generate a set of lifted rules \cite{Chen_Hofmann_Klassen_McIlraith}. These lifted rules, given a partial state, inform the planner which action to take. The goal regression they use propogates backwards allowing for multiple rules to be generated. Their planner also incorporates axioms in addition to regression. They use axioms to encode the learned rules in order to help reduce the search space to allow for optimal planning. Their new planner was able to significantly improve upon state-of-the-art planners in many paradigms: satisficing, optimal, and numeric.



Regression can be used to enable execution monitoring as seen in both \citeauthor{Fritz_McIlraith} (\citeyear{Fritz_McIlraith}) and \citeauthor{muise2011monitoring} (\citeyear{muise2011monitoring}). The former focuses on producing robust plans that are capable of handling uncertainty. Sensors, motors, and observations can all be imperfect, causing the perception of the state to differ. They introduce an execution monitoring system that aims to avoid needless replanning. Using regression, they can determine whether or not a plan can still achieve the goal, and that can inform the decision on whether further action is needed. \citeauthor{muise2011monitoring} (\citeyear{muise2011monitoring}) focus on exploiting partial order plans (POPs) in their execution monitoring framework, also in imperfect and variable worlds. By approximating the regression of the POP, one can produce a policy that allows an execution monitoring framework to exploit this information and improve the robustness of plan execution.

For work regarding axioms, some research has been done to make axioms more applicable in planning. \citeauthor{speck_2019} (\citeyear{speck_2019}) incorporate axioms natively into symbolic search algorithms. They analyze three methods for encoding axioms: Action-based, Variable-based, and Symbolic Compilation.

Axioms have also been used in domain construction to great success. \citeauthor{Baier_McIlraith} (\citeyear{Baier_McIlraith}) use axioms to help translate temporally extended goals into classical planning. In a more granular sense, they use axioms to help represent parameterized nondeterministic finite automata. Their implementation was more concise, expressive, and solved faster than other methods. \citeauthor{Giacomo_Favorito_Fuggitti_2022} (\citeyear{Giacomo_Favorito_Fuggitti_2022}) use axioms to help reduce Pure-Past Linear Temporal Logic goals into classical and FOND planning. They find that axioms provide an elegant solution to their model. Lastly, \citeauthor{Ivankovic2015OptimalPW} (\citeyear{Ivankovic2015OptimalPW}) demonstrate how axioms can help improve the expressivity of a problem and allow more concise definitions. They demonstrate this on multiple domains and conclude that the resultant domains are more natural and more efficiently searchable.




\label{sec:rw}

\section{Preliminaries}

\begin{definition}[States]
 A \textit{state} $\state$ is a function that maps all variables in a set to a defined value, and a partial state $\partialstate$ is a partial function that maps some variables in set to a value. An assignment is a tuple $(v, d)$ where $v$ is the variable and $d$ is the value the variable has. Any variables without an explicit assignment in a partial state are assumed to have the value $\bot$, representing that it is undefined. As partial states do not exclude the possibility of a complete assignment, all states are also partial states. $\stateSet$ is the set of all possible complete states, and $\partialStateSet$ is the set of all possible partial states. We use the notation $v \in s$ to be shorthand for $v \in \{v' \in \variableSet | s(v) \neq \bot\}$.
\end{definition}

We focus on an extension of fully-observable deterministic (FOD) planning problems to handle axioms. Our definition follows the work by Speck et al \cite{speck_2019} and Helmert \cite{Helmert_2006}, extended where necessary.


\begin{definition}[FOD with Axioms]

  A fully observable deterministic planning problem with axioms is defined as $\foda$:

$\variableSet$ is a finite set of state variables. This set can be partitioned into two distinct sets $\basicVariableSet$ and $\derivedVariableSet$, where $\basicVariableSet$ contains only basic variables (variables affected by operators directly) and $\derivedVariableSet$ contains only derived variables (variables whose values are determined by axioms). Each variable $\basicvar \in \basicVariableSet$ has a corresponding finite domain $\basicvarDomain$. Following this, each variable $\derivedvar \in \derivedVariableSet$ has a corresponding domain $\derivedvarDomain$. Derived variables can only take a value of true or false.

    A state is said to be extended if $\forall v \ in \derivedVariableSet, s(v) \neq \bot$. A complete state is one in which all variables in $\basicVariableSet$ are defined.
      
     $\axiomSet$ is the set of axioms in the problem. Each axiom $\axiom \in \axiomSet$ takes the form of the tuple $\axiomstuple$. $\cond{\axiom}$ is a partial state that represents the conditions that must hold for an axiom to be applicable over a given state. This is also known as the body of the axiom. $\derivedvar$ is the affected derived variable. This is also known as the head of the axiom and is the derived variable that is changed by the application of an axiom.

       $A$ is typically stratified to allow for well-understood semantics \cite{Thiébaux_Hoffmann_Nebel_2005}. This stratification produces an ordering $A_{1}, A_{2}, ..., A_{k}$ and is based on the stratification of the derived variables. For each derived variable $\derivedvar$, we can partition them into sets $D_{1}, D_{2}, ..., D_{k}$ such that the following two rules hold. 
       
       (1) If there exists an axiom $\axiom$ whose head is $d_{i}$ and whose body contains the assignment $d_{j} = 1$, where $d_{j} \in \derivedVariableSet$, then $d_{i} \in D_{i}$, $d_{j} \in D_{j}$ and $j \leq i$. 

       (2) If there exists an axiom $\axiom$ whose head is $d_{i}$ and whose body contains the assignment $d_{j} = 0$, where $d_{j} \in \derivedVariableSet$, then $d_{i} \in D_{i}$, $d_{j} \in D_{j}$ and $j < i$. 

       Following this, axioms are stratified such that an axiom is in $A_{i}$ if the head of an axiom, $\derivedvar$, is in $D_{i}$.
       
       $\operatorSet$ is the set of operators in the problem. Each operator $\operator$ takes the form of a tuple $\operatortuple$. $\pre{\operator}$ is a partial state that represents the set of preconditions, while $\eff{\operator}$ is a set of tuples, or assignments, of the form $\efftuple$ where $v$ is a variable in $\basicVariableSet$ and $d$ is a value in the domain of $\basicvarDomain$. 
      
      $\initial$ is the initial state of the problem. This state is an extended state. $\goal$ is a partial state that represents the goal of the planning problem

\end{definition}


\begin{definition}[Logical Entailment]

  Given two logical formulae $A$ and $B$, logical entailment (denoted by $\models$) states that if $A \models B$ then for any model interpretation of $A$ and $B$ there is no case in which $B$ is false and $A$ is true. In other words, if $A$ is satisfied, then $B$ is also satisfied.

  Any partial state can be treated as a logical formula representing a conjunction of the variable assignments. Given two partial states $p1$ and $p2$, if $p1$ entails $p2$ then any variable with an assignment in $p2$ must have the same assignment in $p1$. $\ps{1} \: \models \ps{2}$ iff $\forall v \in V, (\ps{2}(v) \neq \bot) \rightarrow (\ps{1}(v) = \ps{2}(v))$.

\end{definition}

\begin{definition}[Applicability]
    
An operator $\operator$ is applicable in partial state $\partialstate$ if $\partialstate \models \pre{\operator}$. Applying an operator to a partial state produces a new partial state $\partialstateprime$. This partial state is constructed such that for all variables $\basicvar \in \variableSet$, if an assignment $\efftuple$ exists in $\eff{\operator}$ then $\partialstateprime(\basicvar) = d$ else $\partialstateprime(\basicvar) = \partialstate(\basicvar)$. Let the application of an operator $o$ be denoted as $o(s)$

Similarly, a single axiom $\axiomstuple$ is applicable in a partial state $\partialstate$ if $\partialstate \models \cond{\axiom}$. Applying an axiom to a partial state also produces a new partial state $\partialstateprime$. This new partial state has the assignments $\partialstateprime(\basicvar) = 1$ for $\derivedvar \in \axiomstuple$, and $\partialstateprime(\basicvar) = \partialstate(\basicvar)$ otherwise. 
\end{definition}
Axioms are seldom evaluated in isolation. Instead, the entire set of axioms is evaluated. This is done by first assigning all derived variables the value false. Starting in the first stratum, all applicable axioms are applied to the partial state repeatedly until a fixed point is reached. Upon this happening, the next stratum of axioms are iteratively applied. This continues until no application of an axiom will change the state. Axioms follow negation-as-failure semantics, meaning that if the partial state does not entail the body of an axiom, then the derived variable must be false as a consequence. There are many ways to handle the calculation of axioms, with \citeauthor{speck_2019} \cite{speck_2019} covering various interpretations. Let $A(s)$ denote the evaluation of axioms on a partial state producing an extended partial state $p$. Axioms are ecaluated in respect to a given partial state after each time an operator is applied.

\begin{definition}[Progression]
    The progression of a partial state with an operator $\operator$ is the extended partial state produced after applying an operator, and then evaluating the axioms on the resulting partial state. $\prog = A(o(p))$.
\end{definition}

A plan $\pi = [ \mathit{o_{1}}, \mathit{o_{2}}, ..., \mathit{o_{n}} ]$ is a sequence of operators which, when applied sequentially to an initial state, produces an extended state that entails the goal condition $\goal$. That is, the plan $\pi$, applied in the manner $Prog(s_{0}, a_{1}) = s_{1}$, $Prog(s_{1}, a_{2}) = s_{2}$, ..., produces states $S_{*} = [s_{1}, s_{2}, s_{3}, ..., s_{n}$], where $s_{n} \:\models \goal$.

With the core definitions in hand, we now restate logical regression using our notation (this follows the common definition with respect to \citeauthor{reiter2001knowledge} (\citeyear{reiter2001knowledge})).

\begin{definition}[Logical Regression]
The general regression of a formula $\phi$ over an action $\operator$ is an arbitrary logical formula that defines what must hold prior to $\operator$, such that $\phi$ holds after executing $\operator$. Using $\prog$, letting $\phi$ be an arbitrary formula over $V_{b}$, the general regression $Regr(\phi, o)$ is defined such that:

\[ 
\forall \partialstate \in \partialStateSet, \prog \models \phi \: \text{iff} \: \partialstate \models Regr(\phi, \operator).
\] 

We define logical regression over $V_{b}$ and not $V$ without loss of generality as any formula that contains derived variables can be compiled into one that only uses basic variables.

 
\end{definition}

\section{Approximating Regression with Axioms}
\label{sec:reg}


Approximate regression is a weakened version of regression. It provides a sufficient condition, but does not guarantee that it contains the necessary conditions. Our model of regression approximation follows a similar strategy to \citeauthor{Muise_McIlraith_Belle_2014} (\citeyear{Muise_McIlraith_Belle_2014}). They use Pre-Image Filtering (PRIMF), which restricts regression to produce a partial state, such that when an action is applied, it satisfies $\phi$. PRIMF works with problems with conditional effects. This is done by calculating a set of variables called $Support$, which is the union of the preconditions and variables which influence the conditions an action uses in its conditional effects. We similarly treat $Support$ as the preconditions and variables that may influence the evaluation of axioms. 

To maintain the partial state representation of regressed conditions, we also rely on having a complete state available for the computation. This formula takes a state $\state$ and an action $\operator$. It also takes an extra partial state, $\partialstate$, such that the action $\operator$ when applied to $\state$ takes you to a third state which entails $\partialstate$. In this formula, $\state$ would be called a context state. While PRIMF primarily uses this context state to understand which conditional effects are used, we use it to ensure that derived variables are handled correctly. 

\begin{definition}[Approximate Regression]
    The approximate regression of an action is an arbitrary formula that defines the sufficient conditions of what must hold prior to an action, such that a condition is satisfied after executing said action. Using $\prog$, letting $\phi$ be an arbitrary formula over $V_{b}$, the approximate regression $\regrapp$ is defined such that:
    \setlength{\parskip}{0pt}
    \[\forall p \in P, \: \text{if} \: p \models \regrapp \: \text{then} \: \prog \models \phi\]

\end{definition}


We will now introduce our representation of $Support$.

    \begin{definition}[Support]
    
    Given a state $\state$ and an operator $\operator$ that is applicable in $\state$, $\unt$ is a partial state that only contains assignments to basic variables such that $\state \models \unt$ and adheres to the following property:
    
    
    \begin{align*}
    \forall \derivedvar \in \derivedVariableSet \cap \pre{\operator}, \: A(\unt)(\derivedvar) = \pre{\operator}(\derivedvar)
    \end{align*}
    
This definition states that $\unt$, when axioms are applied to it, contains all necessary assignments to allow an operator to be applicable. One interpretation of this is that $\unt$ is a set of basic variables that are sufficient to ensure that the derived variables needed to apply an operator hold in $s$. $\unt$ is a non-unique set of basic variables. There are many different partial states that satisfy the above condition. In the following sections, we will explore how to find different instantiations of $\unt$ and evaluate them.

 \end{definition}

This leads us to the following definition of our version of approximate regression, which we will call \textit{Axiom Aware Approximate Regression}:

\begin{definition} (Axiom Aware Approximate Regression ($\mathit{AAAR}$))
  Given a complete state $\state_{1}$, a partial state $\partialstate_{1}$, and an action $\operator \in \operatorSet$ such that $Prog(\state_{1}, \operator)$ produces $\state_{2} \models \partialstate_{1}$, we can produce a partial state $\mathit{AAAR}(\partialstate_{1}, \operator, \state_{1})$, which approximates regression, as follows:

\begin{align*}   \mathit{AAAR}(\partialstate_{1}, \operator, s_{1}) = \\\begin{cases} 
      s_{1}(v) & \text{if} 
      \:\pre{\operator}(v) \neq \bot$ and $v \notin \derivedVariableSet\\
      \untparam{s_{1}}{\operator}(v) & \text{else if} 
      \: \untparam{s_{1}}{\operator}(v) \neq \bot\\
      \bot  &  \text{else if} \: \eff{o}(v) \neq \bot\\ 
      \bot  &  \text{else if} \: v \in \derivedVariableSet\\ 
      \partialstate_{1}(v) & \text{otherwise}\\ 
   \end{cases}
\end{align*}

In this definition, $\state_{1}$ is called the \textit{context}. This context state is a known state in which operator $\operator$ is applicable. In our partial state $\mathit{AAAR}(\partialstate_{1}, \operator, \state_{1})$, we want to use the context to determine which basic variables and their corresponding values are required to ensure the precondition is satisfied. $\untparam{s_{1}}{\operator}$ is used to determine which variables and values are required to ensure that the derived variables required are still derived appropriately. Lines 1 and 2 could be combined, but are separated out for clarity. Line 3 undefines all variables whose values are assigned by the application of an action. Line 4 undefines all derived variables. Lastly, for any variables not mentioned before, their values are given by $\partialstate_{1}$. In this case, $\partialstate_{1}$ acts as the formula we wish to have satisfied after calculation.

The main objective of this regression formula is to produce a partial state that is sufficient to entail the formula used in the full unapproximated regression. In our use case, this formula represents the values a state must have for a plan to be executable and to achieve the goal if this action were taken. We can guarantee this if we use a plan $\pi$ to inform our choices of $\state_{1}$ and $\partialstate_{1}$. If we let $\partialstate_{1} = \goal$, with $\operator$ being the last action of $\pi$, and $\state_{1}$ being the final state in produced by the execution of the plan, we can approximate regression where any state that entails $\text{AAAR}(\partialstate_{1}, \operator, \state_{1})$ will be sufficient to achieve the goal. This same argument works inductively for every preceding action in $\pi$, using the previous $\text{AAAR}(\partialstate_{1}, \operator, s_{1})$ as $\partialstate_{1}$ and the previous state produced by executing the plan as the context, repeated until we reach the initial state. This sequence of partial states is called $P_{*}$

This also means that the context should entail the approximated regression and the following propositions hold.

\begin{equation}
    s_{1} \models \text{AAAR}(\partialstate_{1}, \operator, s_{1})\\
\end{equation}
\begin{equation}
    \text{AAAR}(\partialstate_{1}, \operator, s_{1}) \models \text{Regr}(\partialstate_{1}, \operator)\\
\end{equation}

\end{definition}

Lastly, we want to be able to talk about how different instantiations of $\unt$ and different regressed partial states compare. To do this, we introduce \textit{generality}.

\begin{definition}[Generality]
 Let $\mathit{undef}(\partialstate)$ be a function that returns a set of variables in a partial state which are equal to $\bot$, that is, all undefined variables. A partial state $\partialstate_{2}$ is more \textit{general} than another partial state $\partialstate_{1}$ if and only if the following constraints hold: 
 \begin{align}
\partialstate_{1} \: \models \partialstate_{2} \\
|\mathit{undef}(\partialstate_{1})| < |\mathit{undef}(\partialstate_{2})|. 
\end{align}

Likewise, a \textit{most general partial state} $\partialstate_{*}$ is defined as:
\begin{multline}
\forall \partialstate_{x} \in \partialStateSet \: s.t. \: \partialstate_{x} \: \models \partialstate_{*}, \: |\mathit{undef}(\partialstate_{x})| \leq |\mathit{undef}(\partialstate_{*})|
\end{multline}

Conditions (3)-(4) state that for a partial state to be more general than another partial state, it must be (3) entailed by the original state and (4) have more undefined variables in comparison to the other partial state. The most general partial state is not necessarily unique, leading to different states that satisfy the constraints.


\end{definition}

Two last considerations must be taken. The goal state may contain derived variables, and if it does, it cannot be used as the partial state in our regression formula. Therefore, we need to find a way to convert the goal into a usable partial state with only basic variables. To do this, we can define a special goal-achieving action. The action has the preconditions that we are in a goal state and the effect that the goal is achieved. We regress the goal state over this goal-achieving action to produce a state that is useful for the rest of the regression calculations. Secondly, some instantiation methods make additional assumptions. One of our methods does not work with stratified axioms due to the requirement of needing to know if a derived variable is false. This restriction is a soft restriction, as it is possible to convert the problem into one without stratification \cite{roger2024negated}, but it is mentioned nonetheless.

\subsection{Naive Method}



The first version of $\unt$ is a naive method: $\unt(v) = s(v) \: \forall v \in \basicVariableSet$. This method ensures the same derived variables will be assigned through the application of axioms. This method produces the least general regressed states and likely contains redundant or irrelevant information, making it an ideal baseline.

\subsection{Relevant Variables Method}

\begin{algorithm}[t]
\caption{Find Relevant Variables: $\mathit{find}$}\label{alg:find}
\hspace*{\algorithmicindent}\textbf{Input:} Axioms $axiomSet$, Relevant Derived Variables $DV$ (The initial $DV$ will be given by the preconditions of action $\operator$)\\
\hspace*{\algorithmicindent}\textbf{Output:} Relevant Basic Variables $RV$\\
\begin{algorithmic}[1]

\State $RV = \emptyset$
\For{$\axiomstuple$ in $\axiomSet$ such that $d \in DV$}
\For{$v \in \variableSet$ such that $\cond{a}(v) \neq \bot$}
    \If{$v \in \derivedVariableSet$}
        \State $RV \gets RV \cup \mathit{find}(A, \{v\})$
    \Else
        \State $RV \gets RV \cup \lbrace v \rbrace$
    \EndIf
    \EndFor
    \EndFor
\State return $RV$
\end{algorithmic}
\end{algorithm}

The second method is to only have assignments in $\unt$ for variables deemed to be relevant. These are variables which are directly (or indirectly) used in the calculation of relevant derived variables. This may be calculated recursively, as seen in Algorithm \ref{alg:find}, which returns a set of variables whose assignments must be defined in $\unt$. We first find the set of variables that are referenced by the required axioms, axioms which are mentioned in the preconditions of the action we wish to approximate regression with. Basic variables in the precondition are collected in a set, while derived variables have their referenced variables in their respective axioms calculated. This is repeated until only basic variables are returned. The end result is a set of all basic variables which influence the derivation of the required axioms. We can use this set to assign these variables in $\unt$ from the context. This algorithm is likely to generate a more general partial state than the naive instantiation, but not the most general. 


\subsection{Post Processing Method}
We can find alternative approximate regressions by improving the $\unt$ found with the relevant variables method using a technique we refer to as \textit{post-processing}.  Afterwards, we perform a local search algorithm to find more general partial states. We perform this search by choosing variables systematically to undefine in the our approximated partial state. Algorithm \ref{alg:relevant} details this process.


This algorithm uses an additional parameter (heuristic). This heuristic informs the order in which we select variables. We explored different heuristics for ordering, some arbitrary and others based on the domain size. While it can make a significant difference when using constructed examples, in practice this decision did not produce different results. Therefore, we use the order given by the planner for all subsequent results, called the ``in-order'' heuristic.

This instantiation has additional restrictions that the previous methods do not. In order to employ this method, we must restrict the problem to not include preconditions of actions, or conditions of axioms which contain negated derived variables. This is to say, the stratification of axioms must only contain one stratum. This method implicitly selects specific axioms which must fire in order for the formula to be satisfied. To prove that a derived variable is false we have to be able to prove that every axiom which could theoretically make it true does not have the conditions satisfied. This is due to the closed world philosophy \cite{Thiébaux_Hoffmann_Nebel_2005}. Therefore, by only checking a specific subset of axioms through a subset of basic variables in the context, we are not able to conclusively say that a derived variable is false. The previous two methods are able to handle negated derived variables as every axiom is considered.  

\begin{algorithm}[t]
\caption{Post Process Support} \label{alg:relevant}
\hspace*{\algorithmicindent}\textbf{Input:} Axioms $\axiomSet$, Heuristic $h$, operator $\operator$, context $\state$, partial state $\partialstate$\\
\hspace*{\algorithmicindent}\textbf{Output:} Support: $\unt$\\
\begin{algorithmic}[1]

\State
\State $\unt = find(Axioms, pre_{o} \cap \derivedVariableSet$ )
\State $r = AAAR(\ps, \operator, \state)$
\State Sort variables in $\variableSet$ with heuristic h
\For{$v \in \variableSet$ if $v \in \unt, r(v) \neq \bot$}
\State $temp = r(v)$
\State $r(v) = \bot$
\If{$Prog(r, \operator) \models \partialstate$}
\State $\unt(v) = \bot$
\Else
\State $r(v) = temp$
\EndIf
\EndFor
\State Return $\unt$
\end{algorithmic}
\end{algorithm}

The next step would be to exhaustively search for the best approximated regressed state and to compare against full regression. In practice this was too computationally complex, but details can be found in the Appendix.


 
\section{Execution Monitoring}
\label{sec:em}


Execution monitoring is one beneficiary of the ability to produce regressed partial states in the presence of axioms. It is an established testbed with many applications: enabling plan executions to recover under uncertainty, testing the robustness of our partial states, and enabling replanning. It is also a strong tool when attempting to handle generalized planning \cite{srivastava2011new}, having been explored in the area of robotics \cite{fikes1972learning}. In this paper, we use execution monitoring to compare different regression approximation techniques by quantifying how many states they can readily handle.



The framework is generally laid out as follows. First, we begin simulating a computed plan from an initial state. Plans are generated prior to execution. With these plans, we can produce a sequence of regressed states, starting from the goal and iteratively moving through the plan in reverse order: $P_{*}$. At some point in the simulation, we change the value of the state, putting the simulator in a new, unexpected state. Using our regressed partial states, we can check if there is any known partial state that is entailed by our current state. If there is, we know we can recover by continuing the execution of the plan at that spot. If not, we know our partial states do not capture this state, and replanning is necessary. We do not perform any replanning step, but rather demonstrate when it is not necessary. Our method is meant to complement existing execution monitoring frameworks, rather than replace them entirely.


Three core decisions were made when designing our execution monitoring methodology that ultimately determine the probability of states changing. The first is to choose when a change occurs. This is done by pre-selecting a random step in the plan in which our execution will deviate. The second choice is to select the basic variable to change. There are two ways in which we have chosen to select this variable. We use a uniform random method, where every basic variable has an equal chance of being selected. We also use a weighted method that ranks and weighs each choice based on the size of the domain. This method selects variables that can take on more values with high likelihood. The final decision is that we randomly select a new value for our variable to take. Under our approach, the new value must be different from the previous value. If this were not the case, the recovery would be trivial.

Execution monitoring allows us to calculate various metrics to determine how resilient our execution is. One metric we can obtain is the suffix length. Suffix length is the index of the action in a plan (indexed in reverse) in which, if the plan was executed from that point, would result in a state that achieves the goal. Given a partial state $p$, $suffix\_length(p) = argmin_{i}[p \models AAAR(S_{*}[len(S_{*})-i], \pi[len(S_{*})-i], P_{*}[i])$. If no action is entailed then suffix length does not exist. Replanning is not necessary if the changed state entails a partial state obtained through $\mathit{AAAR}$. Another metric we can calculate is the success rate. This measures how many trials result in the changed state entailing a regressed partial state. A high success rate corresponds to a plan that is often able to recover without replanning. Finally, there is the undefined rate. This is the number of undefined variables in all partial states in $P_{*}$, normalized, and is indicative of how much information is generalized, which allows more states to entail our approximate regression.

\section{Evaluation}
\label{sec:res}

We hope to answer the following research questions: 
\begin{enumerate}
  \item How robust does our approximate regression formula make our plans, or in other words, how often are we able to recover without replanning? 
  \item How \textit{general} are our plans? How much redundant information is contained in each state/ in knowing which variables may be undefined?
\end{enumerate}
We hope to answer the first question by calculating the success rate and the suffix length, which is the number of steps to achieve the goal state. High success rates imply robust plans, and suffix length informs the amount of work required to recover. The latter two questions we can analyze by looking at how many basic variables need defined assignments to achieve the goal and how many are set to $\bot$.

All experiments were run on 12th Gen Intel(R) Core(TM) i7-12700H with 32 GB of RAM. The plans were generated in advance or collected from the IPC 2004 results \cite{Edelkamp_Hoffmann_2005}, in the cases of Promela domains. 

\subsection{Domains}


Many domains were found via a public repository\footnote{https://github.com/dosydon/axiom\_benchmarks}, a popular repository for PDDL domains with axioms \cite{dosydon_2017}. We also altered two domains from IPC 23 \cite{ipc23}. Domains were chosen based on availability and the number of problem instances. Some problem sets in \cite{dosydon_2017} were omitted due to having only two instances or not having axioms.

The domains that come from IPC 4 include Promela-Philosophers, Promela-Optical-Telegraphs, Miconic, PSR, Blocksworld, and Sokoban. Each uses axioms in a different way. Philosophers and Optical Telegraphs use axioms to define deadlocks. Miconic and Sokoban axioms are both used to derive reachability. The former refers to a floor being reachable, and the latter refers to a grid position. Blocksworld uses axioms to define properties such as \verb|above| and \verb|holding|. Lastly, PSR uses axioms to represent the transitive flow of electricity.







In addition to these domains, we modify the following domains to include axioms. All of these domains come from IPC 23 \cite{ipc23}.

\noindent \textbf{(Labrynth)} The goal of this domain is to make a plan to guide an agent through a dynamic maze. The maze can shift along each direction in a torus fashion, changing the paths through the maze. We use axioms to introduce reachability into this domain in a similar way to the Sokoban domain.

\noindent \textbf{(Quantum Circuit Layout Synthesis)} The goal of this problem is to first map logical qubits onto physical qubits. We modify this domain such that some variables are no longer given by the actions, but rather intermediate variables are given, and the initial variable must be derived. These axioms are simpler than those in other domains, but are still representative of how axioms may be used.

We gathered plans from each domain using the FastDownward planner \cite{Helmert_2006}. These plans are not necessarily optimal plans, but rather just a satisficing plan. Due to this, we just ran the stock version of FastDownward with no additional configuration; time was irrelevant. Not every instance of each domain has a corresponding plan due to the computation time it takes to produce them.

We ran an execution monitor for 10,000 iterations for every problem instance across four different configurations. The configurations are partitioned via two decision points. The first is what method should be used to instantiate $Support$. V1 (Version 1) and V4 both use the naive method, specified in Section 4.1, to achieve the initial $Support$ with respect to the goal-achieving action. V2 and V3 use the post-processing method in Section 4.3 and the ``in-order'' heuristic. The relevant variables method of instantiation was omitted as post-processing was similar, leading to a marginal difference. The second choice is whether to continue to post-process the $Support$ during subsequent regression calculations. V1 and V2 choose to omit this step, while V3 and V4 do not. These configurations allow us to explore two different axes to apply our approximated regression formula. Domains which had stratified axioms present are unable to use post-processing and therefore have an N/A.

\begin{table}[t]
    \centering
    \begin{tabular}{|c|c|c|c|c|c|}
        \hline
         Domain & Probs & V1 & V2 \& V3 & V4 & Max\\
         \hline
         Blocksworld & 9 & \textbf{12.8} & N/A & N/A & 12.8\\
         \hline
         Labrynth & 24 & 2.3 & \textbf{31.4} & \textbf{31.4} & 99.9\\
         \hline
         Miconic & 50 & 31.1 & \textbf{49.1} & \textbf{49.1} & 73.3 \\
         \hline
         Optical & 25 & 6.8 & \textbf{73.2} & 48.1 & 80\\
         \hline
         Philosophers & 48 & 12.8 & \textbf{35.4} & 27.1 & 50\\
         \hline
         PSR-large & 43 & 0.5 & \textbf{24.9} & \textbf{24.9} & 5.7\\
         \hline
         Quantum & 3 & 12.1 & \textbf{24.6} & \textbf{24.6} & 39.7\\
         \hline
         Sokoban & 6 & \textbf{0} & N/A & N/A & 0\\
         \hline
    \end{tabular}
    \caption{Average percentage of undefined variables (Undefined rate) across various configurations over entire plans. N/A is for domains with stratified axioms. Best method is bolded.}
    \label{tab:tab1}
\end{table}



\subsection{Undefined Rate}
Table \ref{tab:tab1} contains information on how many variables are undefined across an entire plan. For each configuration, a plan is used to generate a series of states. Given this series, we calculate the percentage of variables at each step which could have been undefined and average them. Percentages are used as each instance has a different number of variables. 50 would mean, on average, half of all basic variables are set to $\bot$ across the entire plan execution. The max column shows the greatest reduction given a single step. In every case, V2 and V3 are the same. Consequently, in all domains tested, it is the same to post-process only the goal as opposed to all of the steps. V4 is the third most efficient method. The naive method is, as expected, the least effective method. 

Some of the examples have a low undefined rate. This can be explained by analyzing the number of basic variables in the domain. With domains such as Sokoban, the majority of the variables are derived. For example, the first Sokoban instance has three basic variables and 70 derived variables. This makes it difficult to undefine variables since only zero to three basic variables can be undefined. Conversely, the 6\_6 instance of Labrynth, with 1375 basic variables and 36 derived variables leads to larger improvements.

\begin{figure}[t]
    \centering
    \includegraphics[width=1\linewidth,height=4.5cm]{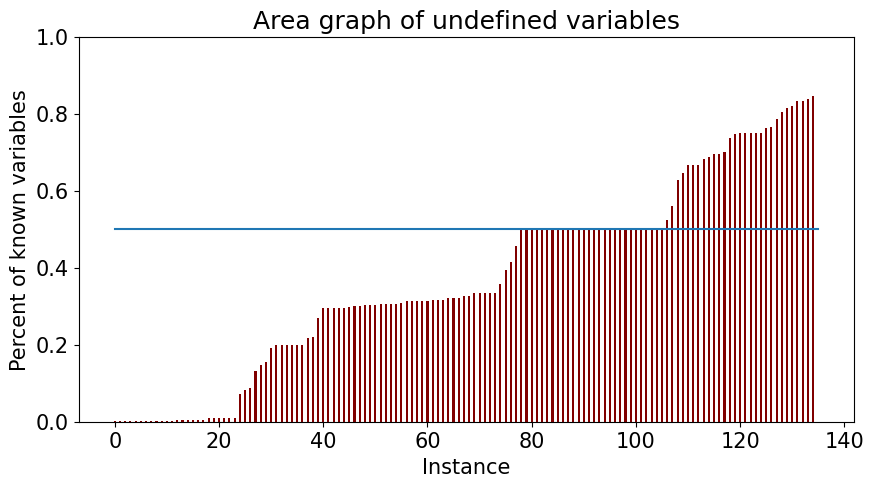}
    \caption{Area graph showing how many undefined variables are in each instance}
    \label{fig:percentageall}
\end{figure}

\subsection{Experimental Configuration}
\begin{figure}[t]
    \centering
    \includegraphics[scale=0.55]{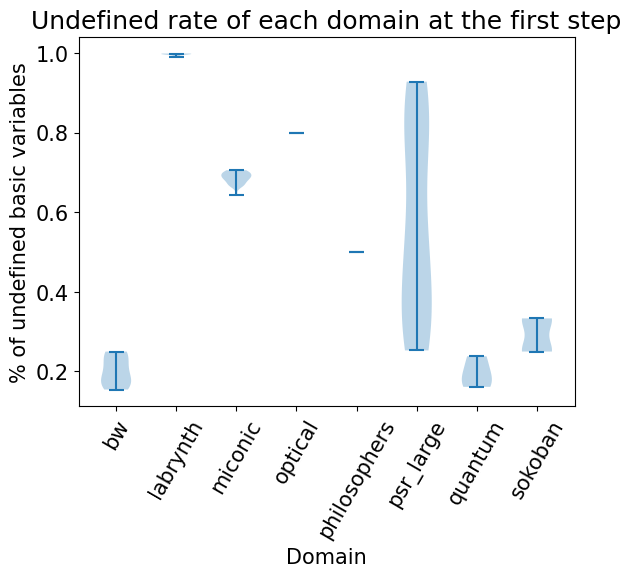}
    \caption{Violin plot of undefined variables}
    \label{fig:violins}
\end{figure}

Figure \ref{fig:percentageall} shows how much information is required for each domain, one step from the goal. This gives a sense of what is required for a partial state to be able to achieve the goal. Over all domains tested, approximately half of all instances are able to undefine half of all basic variables. This graph looks solely at regressing the goal state with the post-processing. Figure \ref{fig:violins} shows that the variables are undefined for each domain one step from the goal by domain for every instance in a domain. Notably, most domains are consistent with how much redundant information exists, with some being exactly the same across all instances: e.g. philosophers always have 50\% of basics undefined. 

\begin{figure}[t]
    \centering
\includegraphics[width=0.85\linewidth]{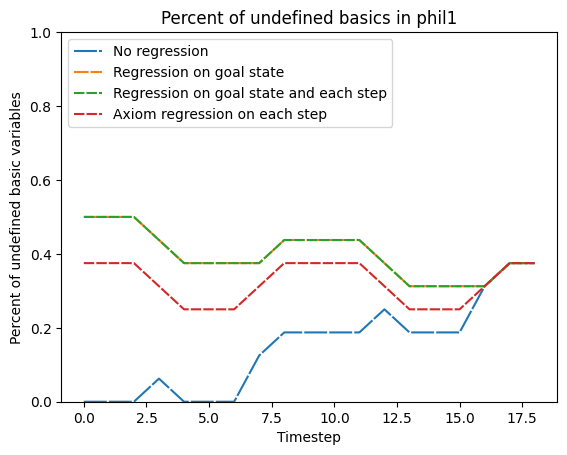}
\caption{Regressing philosophers problem 1}
\label{fig:phil_undef}
\end{figure}
\begin{figure}[t]
    \centering
\includegraphics[scale=0.5]{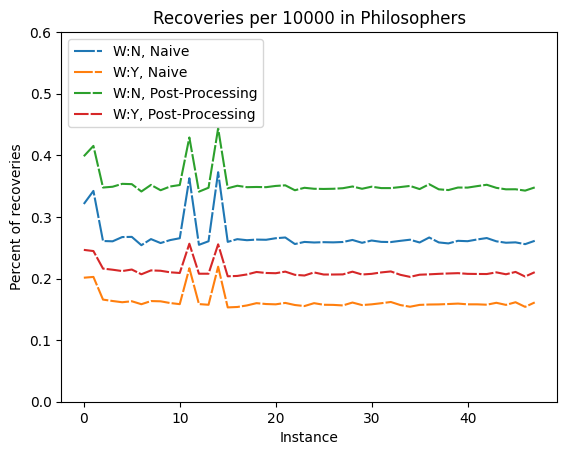}
\caption{The number of recoveries across all instances of Philosophers}
\label{fig:recoveries}
\end{figure}

To further explore how these domains change when regression is applied, we will focus on the Philosophers problems in particular. Figure \ref{fig:phil_undef} shows the first instance of promela-philosophers applied over four configurations. 

\subsection{Execution Monitoring and Recovery}
We can now introduce our execution monitoring and see how our regression approximation can help with execution. The first metric we will analyze is whether the execution could recover without replanning. Looking at one domain, Figure \ref{fig:recoveries} shows the average recoveries for Philosophers. W corresponds to whether the variables were weighted with $Y$ meaning yes and $N$ meaning no, The second value corresponds to whether the naive method was used for the initial $Support$ or the post-processing method. We tested post-processing for subsequent regression calculations, but it had no impact. We can see that the configuration used does play a significant role in whether the monitor recovers. Weighting variables is highly domain-specific, with some domains, such as Optical, performing ~25\% worse on average when using weighted variables. We can look at the recovery rate of all domains, both the largest and smallest instances, in Table \ref{tab:tab3}. This table reports the configuration where post-processing is used at all steps and variables are weighted. For domains with stratified axioms, the relevant variables method was used. The success rate is domain-dependent, but many domains avoided replanning.

The second metric we can analyze is suffix length. Table \ref{tab:tab2} gives numerical data on suffix lengths across the largest instance of each domain. We found that across the board, the configuration has a marginal impact on which step of a plan the execution monitor is able to recover. STD is the standard deviation of when the execution monitor is able to recover. From this, we can see that the execution monitor is able to restart the execution halfway through the plan. While this may seem intuitive, it is possible to construct scenarios where any average suffix is possible.


\begin{table}[t]
    \centering
    \begin{tabular}{|c|c|c|c|}
    \hline
         Instance & RR Small & RR Large & All Conf Same\\
         \hline
         Blocksworld & 24.54\% & 2.21\% & Y\\
         \hline
         Labrynth & 40.75\% & 0.1858\% & Y\\
         \hline
         Miconic & 67.29\% & 48.33\% & Y \\
         \hline
         Phil & 24.65\% & 21.03\% & N \\ 
         \hline
         PSR Large & 71.60\% & 18.87\% & N\\ 
         \hline
         Opt & 40.61\% & 40.71\% & N \\ 
         \hline
         Quantum & 30.04\% & 23.21\% & N\\
         \hline
         Sokoban & 0.25\% & 0.24\% & N \\ 
         \hline
    \end{tabular}
    \caption{Plan recovery rate on the smallest and largest instances}
    \label{tab:tab2}
\end{table}

\begin{table}[t]
    \centering
    \begin{tabular}{|c|c|c|c|}
    \hline
         Instance &  Average Suffix & STD & Plan Length\\
         \hline
         Blocksworld 9 & 18 & 9.3 & 30 \\
         \hline
         Labrynth 6\_6 & 5.6 & 2.2 & 8 \\
         \hline
         Miconic 30\_4 & 25.5 & 17 & 60 \\
         \hline
         Phil 48 & 231.6 & 136.8 & 441 \\ 
         \hline
         PSR large 26 & 15.1 & 9.4 & 31 \\ 
         \hline
         Opt 9 & 71.3 & 40.6 & 140 \\ 
         \hline
         Quantum 3 & 16.4 & 9.9 & 36\\
         \hline
         Sokoban 7 & 15 & 0 & 15 \\ 
         \hline
    \end{tabular}
    \caption{Plan suffix across various instances}
    \label{tab:tab3}
\end{table}



\section{Conclusion}
\label{sec:con}


Axioms are a powerful tool that offer many benefits and can simplify complex phenomena \cite{Thiébaux_Hoffmann_Nebel_2005}. Despite their utility, many extensions and applications of planning do not take advantage of them. We set out to improve this by introducing an approximation for regression over actions with axioms in the domain. We have implemented an approximate regression algorithm for fully observable deterministic domains with axioms. 

Three different versions of this method were created: a naive method which keeps all basic variables to ensure that axiom calculation is performed exactly the same as the state used to approximate regression, a relevant variables method which only keeps the states required for specific axioms, and a post processing method which chooses a specific set of axioms which can achieve the desired result. Additionally, the appendix contains other potential methods.


Using our newly regressed partial states, we introduce an execution monitoring framework. This enables robust behaviour during plan execution by often avoiding replanning if something goes awry. We also use it to test statistical properties of our regression methodology.

Across eight domains, we have identified many key insights. The first is that in current benchmarks, only applying post-processing to the goal state produces an approximate regression over the plan that is just as good as applying post-processing at every step. Problems may be constructed where this is not the case, which raises the question of whether this is a property of real-world domains or if existing benchmarks are simply too limited in scope. Roughly half of the time, across all domains, the execution monitoring system avoided replanning. This allows applications of planning to save computing time. Additionally, our approach was able to greatly generalize plans across all domains, with differences in the number of basic variables between domains having a great impact.



Moving forward, we hope to expand the algorithm to handle both conditional and non-deterministic effects, expand the datasets used, and explore why heuristics did not provide a measurable improvement in practical domains. Another area of future work would be to expand up the benchmarks that contain axioms. To our knowledge there are limited examples. To fully explore the capabilities of such an algorithm, domains will need to be created or modified to include these richer specifications.




\clearpage

\section*{Acknowledgments}
We acknowledge the support of Scotiabank, in conjunction with the Smith School of Business Scotiabank Centre for Analytics \& AI, and Mitacs.
\bibliography{aaai2026}

\clearpage
\section*{Appendix}

\subsection{Full Regression and Exhaustive Search}
\label{contextless}

Given $\basicVariableSet$, one can calculate the Cartesian product of all assignments that generate a partial state which satisfies the definition of $\regrapp$, with $\bot$ being a valid assignment. This Cartesian product represents a version of an unapproximated regression of the system, or the full arbitrary formula $\phi$; the version that we attempt to use when computing full regression. This is done by translating the cartesian product into a disjunction of states, each of which is a conjunction of assignments. We represent full regression in this manner because we need the formula to take the form of partial states. Every possible state which in which an operator is applicable must entail at least one of these partial states. This form of regression contains redundant information but is an upper bound. Calculating the Cartesian product of all assignments is an expensive operation and is only feasible for very small problems in terms of $\basicVariableSet$. We were only able to use it in the smallest problems within reasonable time bounds. Similar to the naive method, this serves as an informative baseline. Despite this it does serve as an excellent comparison tool to evaluate our selected approximated regression.

From this cartesian product we are able to extract individual partial states to be used as approximated regressed states. Each one in the cartesian product is a valid approximate regression. With this said we can introduce our new method, the exhaustive method.

The exhaustive method looks for the partial state that is the most general and still satisfies the condition that given our plan we are able to execute it and achieve the goal. This method takes into consideration the context and ensures that the propositions (1) and (2) in section 4 are satisfied. As every possible initialization is in the cartesian product this is guaranteed to be the best solution.

\end{document}